\documentclass[letterpaper]{article} % DO NOT CHANGE THIS
\usepackage{aaai25}  % DO NOT CHANGE THIS
\usepackage{times}  % DO NOT CHANGE THIS
\usepackage{helvet}  % DO NOT CHANGE THIS
\usepackage{courier}  % DO NOT CHANGE THIS
\usepackage[hyphens]{url}  % DO NOT CHANGE THIS
\usepackage{graphicx} % DO NOT CHANGE THIS
\usepackage{natbib}  % DO NOT CHANGE THIS AND DO NOT ADD ANY OPTIONS TO IT
\usepackage{caption} % DO NOT CHANGE THIS AND DO NOT ADD ANY OPTIONS TO IT
\usepackage{algorithm}
\usepackage{algorithmic}

\usepackage{booktabs}       % professional-quality tables
\usepackage{amsfonts}       % blackboard math symbols
\usepackage{nicefrac}       % compact symbols for 1/2, etc.
\usepackage{microtype}      % microtypography
\usepackage{xcolor}         % colors
\usepackage{amsmath}
\usepackage{amsthm}
\usepackage{multirow,}
\usepackage{colortbl}
\definecolor{lightpastelpurple}{rgb}{0.69, 0.61, 0.85}
\colorlet{LightGreen}{lightpastelpurple!40}
\usepackage{newfloat}
\usepackage{listings}
\DeclareCaptionStyle{ruled}{labelfont=normalfont,labelsep=colon,strut=off} % DO NOT CHANGE THIS
\floatstyle{ruled}
\newfloat{listing}{tb}{lst}{}
\floatname{listing}{Listing}
\title{\textsc{SAVE}: \underline{S}egment \underline{A}udio-\underline{V}isual \underline{E}asy way using Segment Anything Model}
\author {
    Khanh-Binh Nguyen\textsuperscript{\rm 1},
    Chae Jung Park\textsuperscript{\rm 1}
}
\affiliations {
    \textsuperscript{\rm 1}Department of Cancer Big Data and Artificial Intelligence, National Cancer Center, South Korea\\
    \{binhnk,cjp\}@ncc.re.kr
}

\usepackage{bibentry}
\begin{document}

\maketitle

\begin{figure*}
  \centering
    \includegraphics[width=0.8\linewidth, height=0.45\textwidth]{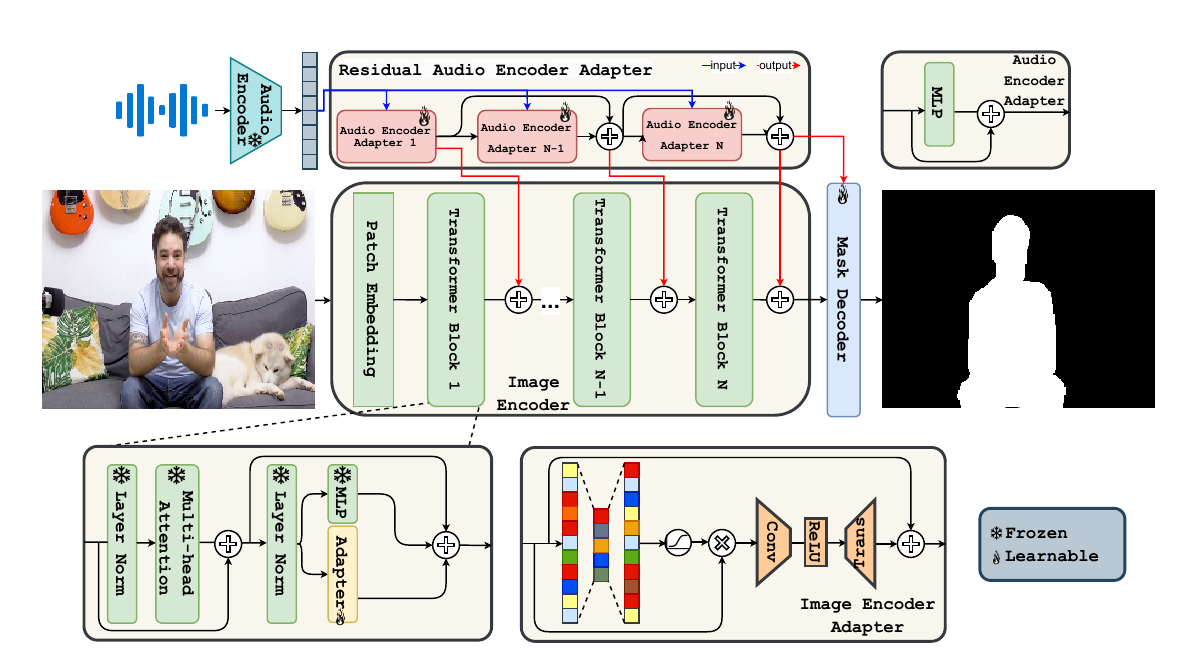}
    \caption{The workflow of SAVE. The image encoder is frozen while an adapter layer is placed into each transformer block to gain expertise specific to the audio-visual domain. The residual audio encoder adapter transforms the audio features and inject them into transformers block for fusing, output of the last residual audio encoder adapter is used as the sparse prompt for mask decoder.}
    \label{fig:framework}
\end{figure*}

\begin{abstract}
The primary aim of Audio-Visual Segmentation (AVS) is to precisely identify and locate auditory elements within visual scenes by accurately predicting segmentation masks at the pixel level.
Achieving this involves comprehensively considering data and model aspects to address this task effectively.
This study presents a lightweight approach, SAVE, which efficiently adapts the pre-trained segment anything model (SAM) to the AVS task.
By incorporating an image encoder adapter into the transformer blocks to better capture the distinct dataset information and proposing a residual audio encoder adapter to encode the audio features as a sparse prompt, our proposed model achieves effective audio-visual fusion and interaction during the encoding stage.
Our proposed method accelerates the training and inference speed by reducing the input resolution from 1024 to 256 pixels while achieving higher performance compared with the previous SOTA.
Extensive experimentation validates our approach, demonstrating that our proposed model outperforms other SOTA methods significantly.
Moreover, leveraging the pre-trained model on synthetic data enhances performance on real AVSBench data, achieving 84.59 mIoU on the S4 (V1S) subset and 70.28 mIoU on the MS3 (V1M) set with only 256 pixels for input images.
This increases up to 86.16 mIoU on the S4 (V1S) and 70.83 mIoU on the MS3 (V1M) with inputs of 1024 pixels.
\end{abstract}


\section{Introduction}
\footnote{Preprint. Under review.}
The task of audio-visual segmentation (AVS) focuses on identifying and delineating auditory elements corresponding to audio cues within video frames.
An ideal AVS model requires dual-level recognition: semantic and instance levels.
However, in practical scenarios, semantic categorization often suffices for localizing sounding objects, achievable by training on artificially constructed data with semantic-consistent image-audio pairs. 
Prior studies have approached this through self-supervised learning, drawing upon audio-visual signals for training objectives \cite{chen2021localizing,hu2020discriminative,hu2021class,lin2023unsupervised,liu2022exploiting,shaharabany2023autosam}.
However, these methods lack precise pixel-level supervision, resulting in coarse segmentation.
This limitation hampers AVS applications in scenarios demanding accurate segmentation, such as video surveillance, multi-modal video editing, and robotics.
\citet{zhou2022audio} recently addressed AVS via supervised learning, manually annotating a video dataset with pixel-level segmentation for sound-related objects.
\citet{duan2024cross} leverages audio and visual modalities as soft prompts to dynamically adjust the parameters of pre-trained models based on the current multi-modal input features.
\citet{chen2023closer} proposes a new cost-effective strategy to build challenging and relatively unbiased high-quality audio-visual segmentation benchmarks.

On the other hand, \citet{mo2023av} attempts to leverage the Segment Anything Model \cite{kirillov2023segment} to perform AVS tasks.
However, the basic SAM \textit{fails to establish semantic connections between audio and images during the pre-training phase for the encoder}.
Even when integrating audio prompts in the decoder phase later on, the fusion of audio-visual features remains superficial, proving inadequate for supporting the AVS task due to the lightweight design of the SAM decoder.
In contrast, \citet{liu2024annotation} suggests a fusion of the visual and audio features by employing an adapter that injects the audio features into each transformer block in the image encoder.
Despite demonstrating some effectiveness, SAMA-AVS \textit{overlooked the significance of the sparse prompt towards the mask decoder of SAM, a crucial element for achieving superior performance}. 
Another notable work is GAVS \cite{wang2023prompting}, which addresses this downside of SAMA-AVS by employing a Semantic-aware Audio Prompt (SAP) to assist the visual foundation model in focusing on sounding objects.
However, it also \textit{neglected the importance of the fusion of visual and audio features}.
Moreover, \cite{mo2023av,liu2024annotation,wang2023prompting} are inefficient methods since they require an input of 1024 pixels with the ViT huge architecture \cite{dosovitskiy2020image}, which induces a large number of computational.

Consequently, this study seeks to overcome these limitations by introducing two innovative modules: 1) an image encoder adapter and 2) a residual audio encoder adapter.
In contrast to the SAMA-AVS adapter, our design employs a residual connection in the residual audio encoder adapter to preserve the information introduced into each transformer block.
Subsequently, our image encoder adapter is designed to enhance the audio-visual fused features in both channel and spatial dimensions.
The output from the final audio residual audio encoder adapter serves as the sparse prompt for SAM's mask decoder, containing the audio-agnostic information infused into the image encoder, thereby significantly improving performance.
Moreover, we introduce SAVE, the first SAM-based AVS model surpassing previous state-of-the-art methods with just 256 pixels for input resolution, achieving groundbreaking results on the AVSBench dataset, which highlights the effectiveness of our proposed method.

In summary, the key contributions include:
1) We propose an image encoder adapter for the SAM model that enhances dataset-specific knowledge transfer for audio-visual fusion across the channel and spatial dimensions.
2) We propose a residual audio encoder adapter for the SAM model that enhances audio features for the audio-visual fusion stage and uses these fused features as a sparse prompt for the mask decoder.
3) We present SAVE, a lightweight SAM-based model for AVS tasks that combines the proposed adapter modules.
By incorporating only 256 pixels for input resolution, SAVE reduces the downside of fine-tuning the SAM-based models and accelerates the inference speed while achieving better performance.

\section{Related Work}
In recent studies, researchers have delved into a spectrum of audio-visual tasks to gain a comprehensive grasp of multimedia resources.
These tasks span audio-visual sound separation \cite{gao2021visualvoice,tzinis2022audioscopev2}, visual sound source localization \cite{chen2021localizing,hu2021class,lin2023unsupervised,liu2022exploiting,qian2020multiple,song2022self}, and audio-visual video understanding \cite{lee2020cross,tian2020unified}.
Visual sound source localization (VSSL), also known as audio-visual segmentation, aims to pinpoint and segment regions of sounding objects based on their audio signals.
Current approaches \cite{chen2021localizing,liu2022exploiting} primarily hinge on the co-occurrence of audio and visual signals, providing weak instance-level supervision, which poses challenges in predicting precise pixel-level segmentation masks.

To address this limitation, \citet{zhou2022audio} introduced a benchmark dataset annotated with masks for sounding objects in video frames, marking a significant step in supervised AVS tasks.
Segment Anything (SAM) \cite{kirillov2023segment}, the pioneering model for image segmentation, was pre-trained on a vast dataset comprising millions of images associated with billions of segmentation masks.
The versatility of SAM extends to various prompts, including points, boxes, masks, and texts, making it applicable across diverse visual problems, such as medical image segmentation \cite{cheng2023sam,gao2023desam,ma2023segment,shaharabany2023autosam,wang2023sam,wu2023medical}, weakly-supervised semantic segmentation \cite{chen2023segment,he2023weakly,jiang2023segment}, few-shot segmentation \cite{liu2023matcher,zhang2023personalize}, 3D vision \cite{chen2021localizing,liu2023segment}, shadow detection \cite{chen2023sam,wang2023detect,zhang2023sam}, and camouflaged object segmentation \cite{chen2023sam,ji2023sam,ji2023segment}.

While some works \cite{mo2023av,liu2024annotation,wang2023prompting} have explored SAM for the AVS task, their outcomes have been less satisfactory.
In this paper, we evaluate the performance of vanilla SAM and introduce SAM-based models specifically designed for AVS tasks.
Adapters, recognized as an efficient method in parameter-efficient transfer learning, introduce minimal parameters to effectively harness knowledge from pre-trained models and apply it to related problems.
They have found widespread use in both natural language processing (NLP) tasks \cite{karimi2021compacter,liu2022few,sung2022lst} and computer vision (CV) tasks \cite{chen2023sam,chen2022vision,li2022exploring,lin2022frozen,liu2023explicit}, and more recently, have been applied to solve multi-modal problems \cite{lee2020parameter,lin2023vision}.

\section{Method}
\subsection{Image Encoder Adapter module}
The most resource-demanding part of SAM is the global update of the image encoder during the fine-tuning process, which leads to significant computational costs.
To infuse knowledge specific to the dataset into the image encoder in a cost-effective manner, we introduce an image encoder adapter layer.
Specifically, during fine-tuning, we freeze all parameters of the original image encoder and add a proposed image encoder adapter to each transformer block, as shown in Figure \ref{fig:framework}.
We modify the image encoder across both the channel and spatial dimensions.
For the channel dimension, we first reduce the resolution of the input feature map to $C\times1\times1$ using global average pooling.
Then, a linear layer is employed to shrink the channel embeddings and another linear layer to expand them, keeping a compression ratio of $0.25$.
Next, we calculate the weights for the channel dimension using a sigmoid function and multiply them with the input feature map to produce the input for the next layer.
For the spatial dimension, we downsample the spatial resolution of the feature map by a factor of two using a convolutional layer and restore the spatial resolution using a transposed convolution, maintaining the same number of channels as the input.
The overall function for the transformer block with the proposed image encoder adapter is defined as:
\begin{equation}
\begin{split}
    x_{i}^{a} &= \text{MHSA}(\text{LN}(x_{i-1})) + x_{i-1} \\
    x_{i} &= \text{MLP}(\text{LN}(x_{i}^{a})) + \text{Adapter}(\text{LN}(x_{i}^{a})) + x_{i}^{a}
\end{split}
\end{equation}
where MHSA, MLP, LN, and Adapter are the Multi-head Self-Attention, Feed-Forward Network, Layer Norm, and our proposed Adapter layer, respectively.
$x$ denotes the input, and $i$ indicates the i-th element.

\subsection{Residual Audio Encoder Adapter module}
The residual audio encoder adapter is a composition of multiple audio encoder adapters, interconnected via a residual connection.
Each audio encoder adapter (PE) consists of a 2-layer MLP and a residual connection.
The quantity of these encoders corresponds to the number of transformer blocks in the image encoder, as we inject the audio features subsequent to each transformer block. 
The output of the residual audio encoder adapter is subsequently utilized as the sparse prompt.
However, due to a misalignment in dimensions, we employ an MLP to map the feature from the image encoder embedding dimension ($C\times H\times W$) into the prompt embedding dimension ($256\times H\times W$). 
Finally, this sparse prompt is input into the mask decoder and fine-tuned with the gradients from the audio-visual fused features following SAM procedure.

\begin{equation}
\begin{split}
    \mathbf{prompt}_{\text{out}} &= \text{PE}_{N}(F_A + \text{PE}_{N-1}(F_A)) \\
    &+ \text{PE}_{N-1}(F_A + \text{PE}_{N-2}(F_A)) + \dots + \text{PE}_{1}(F_A)\\
    \text{PE}_i(F_A) &= \text{MLP}(F_A) + F_A
\end{split}
\end{equation}
where $\mathbf{prompt}_{\text{out}}$ is the sparse prompt to use for the mask decoder and $\text{PE}_{i}$ is the i-th audio encoder adapter.

\subsection{Fine-tuning strategy}
There are several issues of SAM-based models in terms of computational burden when using ViT-H architecture and an input image resolution of $1024\times1024$ pixels.
To ease these drawbacks, we resize the input image to $256\times256$ pixels.
This resizing strategy enables training on low-memory GPUs such as the NVIDIA GeForce RTX 3090, which only has 24GB of memory.
In addition, using lower-resolution input also accelerates the training with a larger batch size and reduces the training as well as inference time.

Specifically, we modify the Relative Position Encoding token from the pre-trained SAM by applying bilinear interpolation to $256\times256$ pixels.
However, such interpolation notably diminishes performance.
Therefore, employing an adapter for the image encoder is the best approach to mitigating this performance degradation.

\subsection{Learning Objectives}
\paragraph{Segmentation Loss.} During the training process of the model, we employ the binary cross-entropy loss, denoted as $BCE(\cdot)$, to quantify the discrepancy between the mask predicted by the model and the actual mask as:
\begin{equation}
\begin{split}
\mathcal{L}_{seg} = \text{BCE}(M_{pred}, M_{gt})
\end{split}
\end{equation}
where $M_{pred}$ and $M_{gt}$ are predicted masks and ground truth masks, respectively.
%% Mathematical formula for semantic loss?  (CJP)
\paragraph{Semantic Loss.} We adopt a simple IoU (Intersection over Union) loss to optimize the IoU between predicted masks and the ground truth masks following \citet{liu2024annotation}.
\paragraph{Total Loss.} The final loss function is calculated as the sum of the two aforementioned losses:
\label{loss_function}
\begin{equation}
\begin{split}
\mathcal{L}_{seg} = \mathcal{L}_{seg} + \mathcal{L}_{IoU}
\end{split}
\end{equation}

\section{Experiments}
\subsection{Datasets}
\label{sec:datasets}

%  \begin{table}[!htb]
%     
%     \centering
%     \label{tab:dataset}
%     \begin{tabular}{|l|c|c|c|c|}
%     %\toprule
%         subsets         & classes   & videos    & train/valid/test  & labeled frames \\
%         \midrule
%         S4 (V1S)        & 23        & 4,932     & 3,452/740/740     & 10,852 \\
%         MS3 (V1M)       & 23        & 424       & 296/64/64         & 2,120 \\
%         V2              & 70        & 6,000     & 4,750/500/750     & 60,000 \\
%         Synthetic       & 46        & -         & 52,609/500/500    & 58,405 \\
%     %\bottomrule
%     \end{tabular}
%     \caption{Dataset summary.}
% \end{table}

\subsubsection{AVSBench dataset}
The AVSBench dataset \cite{zhou2022audio} is a newly launched dataset for audio-visual segmentation, which comes with carefully human-annotated mask annotations.
This dataset is divided into two unique subsets: the semi-supervised Single Sound Source Segmentation (S4) and the fully supervised Multiple Sound Source Segmentation (MS3).

\paragraph{AVSBench-S4 subset.} The AVSBench S4 (V1S) dataset includes videos that contain at most one sounding object.
During the training phase, only the first frame of each video sequence is annotated with a ground-truth segmentation binary mask.
The goal during inference is to segment sounding objects in all frames of the video.
This subset contains 3,452/740/740 videos for the training, validation, and test sets, respectively, amounting to 10,852 annotated frames.

\paragraph{AVSBench-MS3 subset.} On the other hand, the AVSBench MS3 (V1M) dataset is made up of videos that may contain multiple-sounding objects.
All frames in these videos are annotated with masks for both the training and evaluation stages.
This subset has 296/64/64 videos for training, validation, and testing, respectively, totaling 2,120 annotated frames.

\subsubsection{AVSBench-Semantic dataset} 
The AVSBench-Semantic (AVSS) dataset is an enriched dataset by adding a third Semantic-labels subset that provides semantic segmentation maps as labels.
AVSS contains S4 and MS3 subsets, namely V1S and V1M, respectively.
AVSS is enriched in video amount and audio-visual scene categories.
We only use the additional subset, V2, to evaluate our model.
The V2 subset contains 6,000 videos in 70 categories, which are split into 4,750/500/750 for training, validation, and testing, respectively.
For each video, 10 frames are extracted, which creates 60,000 frames in total.

\subsubsection{AVS-Synthetic dataset}
The AVS-Synthetic dataset \cite{liu2024annotation} is introduced, encompassing 62,609 instances of sounding objects across 46 common categories.
The size of the training set is 52,609, and both the validation and test sets contain 5,000 instances.
The process of constructing this dataset utilizes existing image segmentation datasets such as LVIS and Open Images, and audio classification datasets such as VGGSound.
% We show the dataset summary in Table \ref{tab:dataset}.

\subsection{Implementation Details}
\label{sec:implementation}
Our approach is built using PyTorch and is trained on 2 NVIDIA RTX A6000 GPUs.
Except for the resolution of 224 used for the V2 subset, all experiments are conducted using resolutions of 224 and 1024.
% for a resolution of 1024 and on 2 NVIDIA GeForce RTX 3090 GPUs for resolutions of 256 and 224 (only for the V2 subset).
We employ the AdamW optimizer with a starting learning rate of $2\times10e{-4}$ and carry out training over 80 epochs, using cosine decay.
During the training phase, all images are resized to resolutions of $256\times256~(224\times224)$ and $1024\times1024$.
Our strategy for resizing includes zero-padding the edges for images where both the width and height are less than 256 (224) and using bilinear interpolation for resizing images in other scenarios.
The loss function that oversees the mask predictions is a blend of binary cross entropy and IoU loss, as indicated in Section \ref{loss_function}.
The batch size is established at 2 for a resolution of 1024, and 16 for a resolution of 256 and 224 per GPU.

We measure the segmentation quality by employing established segmentation metrics detailed in \cite{zhou2022audio}.
Specifically, we denoted $\mathcal{M_{J}}$ as the mean Intersection-over-Union (mIoU) between ground-truth binary masks and predicted masks.
Additionally, we use $\mathcal{M_{F}}$, referred to as the F-score, representing the harmonic mean of precision and recall.
In both instances, higher values signify superior segmentation performance.

 \begin{table}
    
    \centering
    \begin{tabular}{l|c|c|c|c}
    %\toprule
        Subsets & \multicolumn{2}{c|}{S4 (V1S)}  &  \multicolumn{2}{c}{MS3 (V1M)} \\
        \midrule
        Methods & $\mathcal{M_{J}}$ & $\mathcal{M_{F}}$ & $\mathcal{M_{J}}$ & $\mathcal{M_{F}}$ \\
        \midrule
        SAM                                     & 55.08 & 0.739 & 53.96 & 0.638 \\
        AV-SAM                                  & 40.47 & 0.566 & -     & - \\
        AP-SAM                                  & 69.61 & 0.796 & 51.58 & 0.578 \\
        SAMA-AVS                                & 81.53 & 0.886 & 63.14 & 0.691 \\
        GAVS                                    & 80.06 & 0.902 & 63.70 & \underline{0.774} \\
        \midrule
        \rowcolor{LightGreen}SAVE (256)           & \underline{84.06} & \underline{0.905} & \underline{64.16} & 0.712 \\
        \rowcolor{LightGreen}SAVE (1024)          & \textbf{85.11} & \textbf{0.912} & \textbf{67.01} & \textbf{0.777} \\
                                                & \textcolor{green}{($3.58\uparrow$)} & \textcolor{green}{($0.026\uparrow$)} & \textcolor{green}{($3.31\uparrow$)} & \textcolor{green}{($0.003\uparrow$)}\\   
    %\bottomrule
    \end{tabular}
    \caption{Comparison between different SAM-based methods on the test sets of S4 (V1S) and MS3 (V1M) subsets of the AVSBench dataset \cite{zhou2022audio}. "256" and "1024" denote the resolution of input images. Notably, \cite{kirillov2023segment,mo2023av,liu2024annotation,wang2023prompting} use "1024" for their work. The best results are in bold, and the second-best are underlined. Improvements against the SOTA are in the last row.}
    \label{tab:MS3_S4}
\end{table}

\subsection{Comparison with SAM-based methods}
We have assessed the performance of various methods on AVSBench, which include SAM \cite{kirillov2023segment}, AV-SAM \cite{mo2023av}, AP-SAM \cite{liu2024annotation}, SAMA-AVS \cite{liu2024annotation}, and GAVS \cite{wang2023prompting}.
For the SAM baseline, we utilized the pretrained model weights of the ViT-H SAM model without any additional training and adopted the maximum segmentation evaluation.
The results from AV-SAM \cite{mo2023av}, AP-SAM \cite{liu2024annotation}, SAMA-AVS \cite{liu2024annotation}, and GAVS \cite{wang2023prompting} are cited directly from their original papers.
The results, as displayed in Table \ref{tab:MS3_S4}, reveal that our proposed SAVE method significantly outperforms all the SAM-based methods by a substantial margin across both subsets.
Notably, on the S4 subset, SAVE exceeds the previous state-of-the-art, SAMA-AVS, with an improvement of \textbf{2.53} $\mathcal{M_{J}}$ while only using input images of \textbf{256} pixels and \textbf{3.58} $\mathcal{M_{J}}$ when using 1024 pixels.
This underscores the effectiveness of our proposed image encoder adapter.

 \begin{table}[!htb]
    
    \centering
    \begin{tabular}{l|l|c|c|c|c}
    %\toprule
        &   Subsets & \multicolumn{2}{c|}{S4 (V1S)}  &  \multicolumn{2}{c}{MS3 (V1M)} \\
        \cmidrule{2-6}
        &   Methods & $\mathcal{M_{J}}$ & $\mathcal{M_{F}}$ & $\mathcal{M_{J}}$ & $\mathcal{M_{F}}$ \\
        \midrule
        \multirow{2}{*}{SSL}        & LVS   & 37.94  & 0.510  & 29.45  & 0.330 \\
                                    & MSSL & 44.89  & 0.663  & 26.13  & 0.363 \\
        \midrule
        \multirow{2}{*}{VOS}        & 3DC  & 57.10  & 0.759  & 36.92  & 0.503 \\
                                    & SST   & 66.29  & 0.801  & 42.57  & 0.572 \\
        \midrule
        \multirow{2}{*}{SOD}        & iGAN & 61.69  & 0.778  & 42.89  & 0.544 \\
                                    & LGVT   & 74.94  & 0.873  & 40.71  & 0.593 \\
        \midrule
        \multirow{9}{*}{AVS}        & AVSBench  & 78.74  & 0.879  & 54.00  & 0.645 \\
                                    % & w/ F.T.   & -     & -     & 51.45 & 0.671 \\
                                    & \textsc{ECMVAE}  &   81.74   & 0.901  & 57.84 & 0.708 \\
                                    & AVSegFormer & 82.06 & 0.899 & 58.36 & 0.693  \\
                                    % & AVSegFormer (512) & 83.06 & 0.905 & 61.33 & 0.730  \\
                                    & AVSC & 81.29 & 0.886 & 59.50 & 0.657  \\
                                    & AuTR & 80.40 & 0.891 & 56.20 & 0.672  \\
                                    & GAVS  & 80.06 & 0.902 & 63.70 & 0.774 \\
                                    & AQFormer    & 81.60 & 0.894    & 61.10   & 0.721 \\
                                    & COMBO    & 84.70 & 0.919    & 59.20   & 0.712 \\
        % \midrule
        % \textsc{LAVISH} \cite{lin2023vision}    & - &   80.10   &   -    &   -   &   -   \\
        \midrule
        \multirow{3}{*}{SAMA-AVS}   & w/o F.T.  & 81.53 & 0.886  & 63.14 & 0.691 \\
                                    & w/ F.T.   & -     & -     & 66.30 & 0.730 \\
           & w/ F.T. (Syn.)    & 83.17 & 0.901  & 66.95 & 0.754 \\
        \midrule
        \rowcolor{LightGreen}       & w/o F.T.          & 84.06 & 0.905 & 64.16 & 0.699 \\
        \rowcolor{LightGreen}\multirow{-2}{*}{SAVE (256)}                            & w/ F.T. (Syn.)      & \underline{84.59} & \underline{0.909} & \underline{70.28} & \underline{0.767} \\
        \rowcolor{LightGreen}                            & w/o F.T.         & 85.11 & 0.912 & 67.01 & 0.739 \\
        \rowcolor{LightGreen}\multirow{-2}{*}{SAVE (1024)}                      & w/ F.T. (Syn.)           & \textbf{86.16} & \textbf{0.921} & \textbf{70.83} & \textbf{0.782} \\
                                    % &             & \textcolor{green}{($4.63\uparrow$)} & \textcolor{green}{($0.035\uparrow$)} & \textcolor{green}{($3.33\uparrow$)} & \textcolor{green}{($0.007\uparrow$)}\\   
    %\bottomrule
    \end{tabular}
    \caption{Comparison with the state-of-the-art method and other related audio-visual methods on test sets of two subsets of the AVSBench dataset. “F.T.” refers to fine-tuning based on the pretrained weights on S4 (V1S), while “F.T.(Syn.)” refers to fine-tuning based on the pretrained weights of the S4 (V1S) model on AVS-Synthetic). "256" and "1024" denote the resolution of input images. Notably, \cite{kirillov2023segment,mo2023av,liu2024annotation,wang2023prompting} use "1024" resolution.
    The best results are in bold, and the second-best are underlined.
    % Improvements against SAMA-AVS are in the last row.
    }
    \label{tab:sota_compare}
\end{table}

\subsection{Comparison with other SOTA methods on AVSBench S4 and MS3 subsets}
Our proposed SAVE method has been compared with the latest AVS techniques \cite{ling2023hear,gao2023avsegformer,liu2023audioa,liu2023audiob,mao2023multimodal,zhou2022audio,huang2023discovering}, as well as other related audio-visual methods.
These include sound source localization (SSL) methods such as LVS \cite{chen2021localizing} and MSSL \cite{qian2020multiple}, video object segmentation (VOS) methods like 3DC \cite{mahadevan2020making} and SST \cite{duke2021sstvos}, and salient object detection (SOD) methods such as iGAN \cite{mao2021transformer} and LGVT \cite{zhang2021learning}.

As per the results in Table \ref{tab:sota_compare}, our approach surpasses existing methods across all metrics in both subsets.
For instance, it achieves a score of \textbf{84.06} for the S4 subset and \textbf{64.16} for the MS3 subset on $\mathcal{M_{J}}$.
This performance is notably superior to SAMA-AVS, even though our method uses only \textbf{256 pixels} as input resolutions compared to SAMA-AVS of \textbf{1024} pixels.

We further fine-tuned our models, which were initially trained on the AVS-Synthetic, for the S4 subset and the MS3 subset, following the approach of SAMA-AVS \cite{liu2024annotation}.
Consequently, our method achieved scores of \textbf{86.16/70.83} for the S4/MS3 subset on $\mathcal{M_{J}}$, marking an improvement of \textbf{2.99/3.88} over SAMA-AVS under the same conditions.

This demonstrates that our method can effectively utilize the insights gained from single-sound object segmentation to enhance the performance of multi-sound objects.
Interestingly, SAVE-1024 pretrained on synthetic data performs exceptionally well on the S4 subset for single objects and the MS3 subset for multiple objects.
This suggests that SAVE possesses a strong generalization.

\subsection{Comparison with other SOTA methods on AVSS V2}
As shown in Table \ref{tab:sota_compare_v2}, we evaluate the capabilities of our model on the AVSS V2 subset.
As we have defined in Section \ref{sec:datasets}, the V2 subset has more data as well as many categories, which makes it more difficult than the S4 and MS3 subsets.
For the V2 subset, we resize the input image into $224\times224$ pixels for SAVE and SAMA-AVS \cite{liu2024annotation} for a fair comparison with other methods.
Other training hyperparameters are kept the same as defined in Section \ref{sec:implementation}.

The proposed method, SAVE, achieves the result of \textbf{71.28} for the $\mathcal{M_{J}}$ and \textbf{0.773} for the $\mathcal{M_{F}}$, which significantly outperforms the SOTA method - GAVS \cite{wang2023prompting} by a large margin of \textbf{3.58} $\mathcal{M_{J}}$.
Interestingly, when we apply our fine-tuning strategy with SAMA-AVS \cite{liu2024annotation} for this subset, it can achieve a high performance of 70.48 and 0.770 for $\mathcal{M_{J}}$ and $\mathcal{M_{F}}$, respectively.
This hinders the SAM-based methods, which could benefit from the fine-tuning strategy rather than the high resolution.
This indicates that the fine-tuned model learned specific knowledge of the domain, and fine-tuning with low cost is an effective and feasible method to reduce domain differences.
Based on this observation, we further study the fine-tuning strategy for SAMA-AVS in Section \ref{sec:fine-tuning}.

 \begin{table}     
    
    \centering
    \begin{tabular}{l|l|c|c}
    %\toprule
        \multirow{2}{*}{Methods}    & \multirow{2}{*}{Backbone}  &  \multicolumn{2}{c}{V2} \\
        \cmidrule{3-4}
        & & $\mathcal{M_{J}}$ & $\mathcal{M_{F}}$ \\
        \midrule
        AVSBench   & PVT-v2    & 62.45 & 0.756 \\
        AVSegFormer   & PVT-v2    & 64.34 & 0.759 \\
        GAVS   & ViT-Base  & 67.70 & \underline{0.788} \\
        SAMA-AVS-FT & ViT-Huge    & \underline{70.48}   & 0.770 \\
        \midrule
        \rowcolor{LightGreen}SAVE       & ViT-Huge & \textbf{71.28} & \textbf{0.795} \\
                                    &          & \textcolor{green}{($3.58\uparrow$)} & \textcolor{green}{($0.007\uparrow$)}\\   
    %\bottomrule
    \end{tabular}
    \caption{Performance on AVS-Semantic V2 dataset. All models use VGGish as an audio backbone and 224 as the input image resolution, except 1024 for GAVS. 
    The best results are in bold, and the second-best are underlined. Improvements against the GAVS are in the last row.
    }
    \label{tab:sota_compare_v2}
\end{table}

\subsection{Comparison with other SOTA methods for unseen classes on AVSS V3 subset}
In line with the approach proposed by \citet{wang2023prompting}, we utilize their suggested V3 subset to evaluate the ability of models to generalize to unseen classes for the AVS task.
This evaluation is conducted under four different settings: 0-shot, 1-shot, 3-shot, and 5-shot. For the zero-shot test, objects in the test set are not included in the training or validation stages.
In contrast, for the other settings, $N=[1, 3, 5]$ data samples are selected for each object and incorporated into the training process to facilitate few-shot learning.

As indicated in Table \ref{tab:sota_compare_v3}, our model delivers the highest performance in the 0-shot setting, demonstrating superior generalization capabilities when faced with unseen objects.
Remarkably, the 0-shot performance of SAVE is already on par with the 5-shot performance of GAVS \cite{wang2023prompting}, and it surpasses the best 0-shot result by a significant \textbf{12.74} in $\mathcal{M_{J}}$.
This suggests that our model exhibits a stronger few-shot learning ability. Furthermore, we observe a substantial increase in performance as we incrementally raise the shot number.

\begin{table*}[!htb]
    
    \centering
    \begin{tabular}{l|c|c|c|c|c|c|c|c}
    %\toprule
        \multirow{2}{*}{Method} &   \multicolumn{2}{c|}{0-shot}  &   \multicolumn{2}{c|}{1-shot}  &   \multicolumn{2}{c|}{3-shot}  &   \multicolumn{2}{c}{5-shot} \\
        \cmidrule{2-9}
        & $\mathcal{M_{J}}$ & $\mathcal{M_{F}}$ & $\mathcal{M_{J}}$ & $\mathcal{M_{F}}$ & $\mathcal{M_{J}}$ & $\mathcal{M_{F}}$ & $\mathcal{M_{J}}$ & $\mathcal{M_{F}}$\\
        \midrule
        AVSBench           & 53.00 & 0.707 & 56.11 & 0.754 & 63.22 & 0.767 & 63.87 & 0.783 \\
        AVSegFormer   & 54.26 & 0.715 & 58.30 & 0.764 & 64.19 & 0.774 & 65.17 & 0.785 \\
        GAVS           & \underline{54.71} & \underline{0.722} & \underline{62.89} & \underline{0.768} & \underline{66.28} & \underline{0.774} & \underline{67.75} & \underline{0.795} \\
        \midrule
        \rowcolor{LightGreen}SAVE           & \textbf{67.45} & \textbf{0.731} & \textbf{71.50} & \textbf{0.770} & \textbf{74.61} & \textbf{0.786} & \textbf{75.35} & \textbf{0.795} \\
                                                & \textcolor{green}{($12.74\uparrow$)} & \textcolor{green}{($0.009\uparrow$)} & \textcolor{green}{($8.61\uparrow$)} & \textcolor{green}{($0.02\uparrow$)} & \textcolor{green}{($8.33\uparrow$)} & \textcolor{green}{($0.012\uparrow$)} & \textcolor{green}{($7.6\uparrow$)} & \textcolor{black}{($0.0\uparrow$)}\\   
    %\bottomrule
    \end{tabular}
    \caption{Performance on AVS-Semantic V3 unseen dataset. All models use VGGish as an audio backbone and 224 as the input image resolution, except 1024 for GAVS. 
    The best results are in bold, and the second-best are underlined. Improvements against the GAVS are in the last row.
    }
    \label{tab:sota_compare_v3}
\end{table*}

\section{Ablation study}
\begin{table*}[!htb]
    \centering
    \begin{tabular}{l|l|cccccccccc}
    %\toprule
        Method & Trained with & ambulance & cat & dog & bus & horse & lion & bird & guitar & piano & computer\_keyboard\\
        \midrule
        \multirow{2}{*}{SAMA-AVS} & Real & 73.63 & 83.87 & 84.88 & 85.09 & 81.21 & 91.28 & 85.64 & 85.78 & 82.91 & 85.99\\
        & Synthetic & 55.32 & 78.81 & 78.66 & 82.70 & 73.88 & 88.51 & 81.99 & 75.11 & 32.26 & 74.01\\
        \midrule
        \multirow{3}{*}{SAVE (256)} & Real & \textbf{75.55} & 89.98 & \textbf{87.75} & \textbf{88.26} & \textbf{82.80} & \textbf{91.37} & \textbf{88.36} & \textbf{89.36} & \textbf{89.12} & \textbf{90.32}\\
        & Synthetic & 67.90 & \textbf{91.37} & 86.15 & 83.96 & 77.94 & 89.64 & 84.98 & 85.31 & 38.95 & 77.08 \\
        \cmidrule{2-12}
        & Real + Syn & 77.75 & 92.33 & 88.97 & 87.65 & 84.36 & 91.78 & 87.46 & 89.64 & 85.64 & 90.73 \\
    %\bottomrule
    \end{tabular}
    \caption{The results are categorized based on direct zero-shot evaluations of models that were trained using synthetic data and then tested on the real test split of AVSBench (S4) \cite{zhou2022audio}. In this context, the "Real" row indicates that the model was trained using the real data from the AVSBench training split. Conversely, the "Synthetic" row signifies that the model was trained using AVS-Synthetic. 
    % The metric used for these evaluations is $\mathcal{M_{J}}$, which stands for mean Intersection over Union (mIoU).
    }
    \label{tab:zero-shot-1}
\end{table*}

\subsection{Zero-shot results on S4 subset}
In this section, we evaluate the pretrained model using synthetic data on the S4 subset \cite{zhou2022audio}.
Table \ref{tab:zero-shot-1} quantitatively showcases the results for the overlapping categories between AVS-Synthetic and AVSBench.
As can be seen, SAVE achieves remarkable results across the categories by a large gap compared with SAMA-AVS under the same settings.
Especially in certain categories, such as "ambulance", "cat", "dog", "horse", and "guitar", SAVE exceeds SAMA-AVS when only using the synthetic data for training, which demonstrates the effectiveness of SAVE for zero-shot transfer to real data.
Furthermore, the model trained with both real and synthetic data outmatches the others across almost all categories.

Furthermore, the results presented in Table \ref{tab:zero-shot-2} compare the performance of SAVE and SAMA-AVS under zero-shot and full-shot training on the S4 subset, with or without pre-training on synthetic data.
Although the performance of SAVE is lower when both pre-trained synthetic data and 0\% real data are used, this testing scheme (without training) does not reflect real-world applications.
This is due to the additional learnable image encoder adapter, which requires training to obtain the information from the data.

Nevertheless, SAVE consistently delivers high performance, whether pre-trained on synthetic data or not.
Notably, the model that is solely trained on AVS-Synthetic achieves a mean Intersection over Union (mIoU) of \textbf{69.85} ($\mathcal{M_{J}}$) without the use of any real data.
When 100\% real data is used, SAVE enhances its performance by \textbf{3.53} $\mathcal{M_{J}}$ and \textbf{1.99} $\mathcal{M_{J}}$, achieving scores of \textbf{85.16} $\mathcal{M_{J}}$ and 84.59 $\mathcal{M_{J}}$ with and without the weights of the AVS-Synthetic pre-trained model, respectively.

 \begin{table*}[!htb]     
    \centering
    \begin{tabular}{l|l|c|c|c|c}
    %\toprule
        \multirow{2}{*}{Real data} & \multirow{2}{*}{Method} & \multicolumn{2}{c|}{w/o P.T}  &  \multicolumn{2}{c}{w/ P.T} \\
        \cmidrule{3-6}
        & & $\mathcal{M_{J}}$ & $\mathcal{M_{F}}$ & $\mathcal{M_{J}}$ & $\mathcal{M_{F}}$ \\
        \midrule
        \multirow{3}{*}{$0\%$} & SAMA-AVS (1024)         & 17.65 & 0.227 & 64.05 & 0.744 \\
        & SAVE (256)                                                          & 8.28 & 0.162 & 65.00 & 0.743 \\
        & SAVE (1024)                                                         & 2.10 & 0.06 & \textbf{69.85} & \textbf{0.785} \\
        \midrule
        \multirow{3}{*}{$100\%$} & SAMA-AVS (1024)       & 81.53 & 0.886 & 83.17 & 0.901 \\
        & SAVE (256)                                                          & 83.11 & 0.895 & 84.59 & 0.909 \\
        & SAVE (1024)                                                         & \textbf{84.06} & \textbf{0.905} & \textbf{85.16} &\textbf{0.911} \\
                                                &   & \textcolor{green}{($1.58\uparrow$)} & \textcolor{green}{($0.009\uparrow$)} & \textcolor{green}{($3.87\uparrow$)} & \textcolor{green}{($0.480\uparrow$)}\\   
    %\bottomrule
    \end{tabular}
    \caption{The model is fine-tuned using varying proportions of real training data from the S4 dataset. The term "w/ P.T." denotes a model that has been pre-trained on AVS-Synthetic. The percentages 0\% and 100\% represent the ratios of real data used from the AVSBench S4 training set. On the other hand, "w/o P.T." refers to standard training that does not utilize pre-trained weights.}
    \label{tab:zero-shot-2}
\end{table*}

\subsection{Effectiveness of each module toward the results on MS3 (V1M) and S4 (V1S) subsets}
\label{sec:fine-tuning}
In this section, we experiment with each additional module (image encoder adapter and residual audio encoder adapter) to show the improvement from each of them.
The results are shown in Table \ref{tab:module_compare}.
For a fair comparison, we also fine-tune the SAMA-AVS with 256 input image resolution.
As we can see, inference SAMA-AVS directly with 256 pixels has a significant drop in performance.
Contrasting with it, fine-tuning the mask decoder for the input resolution of 256 shows comparable performance to its 1024 pixels version on S4 and lower performance on the MS3 subset.
This strongly proves that SAM-based methods do not depend heavily on the input resolution to achieve high performance.
With proper fine-tuning, one can have as high performance as using $1024\times1024$ pixels while only using $256\times256$ pixels.

 \begin{table*}[!htb]     
    
    \centering
    \begin{tabular}{l|c|c|c|c}
    %\toprule
        Subsets & \multicolumn{2}{c|}{S4 (V1S)}  &  \multicolumn{2}{c}{MS3 (V1M)} \\
        \midrule
        Methods & $\mathcal{M_{J}}$ & $\mathcal{M_{F}}$ & $\mathcal{M_{J}}$ & $\mathcal{M_{F}}$ \\
        \midrule
        SAMA-AVS (256)      & 13.66 & 0.197 & 29.73 & 0.479 \\
        SAMA-AVS-FT (256)   & 82.49 & 0.897 & 58.95 & 0.646 \\
        SAVE (256) (*)       & 81.68 & 0.887 & 62.94 & 0.696 \\
        \rowcolor{LightGreen}SAVE (256) (**)      & \textbf{83.11} & \textbf{0.895} & \textbf{64.16} & \textbf{0.699} \\
                                                 & \textcolor{green}{($69.45\uparrow$)} & \textcolor{green}{($0.698\uparrow$)} & \textcolor{green}{($34.43\uparrow$)} & \textcolor{green}{($0.220\uparrow$)}\\   
        \midrule
        SAMA-AVS (1024)       & 81.53 & 0.886 & 63.14 & 0.691 \\
        SAVE (1024) (*)                           & 82.43 & 0.890 & 65.54 & 0.710 \\
        \rowcolor{LightGreen}SAVE (1024) (**)     & \textbf{83.11} & \textbf{0.895} & \textbf{67.01} &\textbf{0.739} \\
                                                 & \textcolor{green}{($1.58\uparrow$)} & \textcolor{green}{($0.009\uparrow$)} & \textcolor{green}{($3.87\uparrow$)} & \textcolor{green}{($0.480\uparrow$)}\\   
    %\bottomrule
    \end{tabular}
    \caption{Effectiveness of each module toward the results on MS3 (V1M) and S4 (V1S) subsets. '*' denotes the residual audio encoder adapter; '**' denotes the model with an additional image encoder adapter at the end of each transformer block. $256$ and $1024$ denote the resolution of the input image. 
    The best results are in bold, and the second-best are underlined. Improvements against the SAMA-AVS are in green.
    }
    \label{tab:module_compare}
\end{table*}
% \vspace{-0.35cm}

SAVE, on the other hand, consistently improves performance.
Specifically, SAVE-256 using only a residual audio encoder adapter with a fine-tuning strategy shows much performance compared with SAMA-AVS using 1024 pixels.
With the additional image encoder adapter, the performance outperforms the SAMA-AVS significantly on both subsets.
Interestingly, SAVE-256 performs better on the S4 subset than SAVE-1024, and vice versa, SAVE-1024 provides a much better segmentation mask on the MS3 subset.
This shows that we could consider the trade-off between efficiency and performance when dealing with single/multiple objects.

\subsection{Ablation for memory and training time efficiency}
We benchmark the batch size and inference FPS for both SAVE and SAMA-AVS \cite{liu2024annotation}.
SAVE-256 enables the training with a higher batch size of 16 per GPU as well as increasing the inference FPS to 35, while SAVE-1024 and SAMA-AVS only allow a batch size of 2 and an FPS of 8.
 \begin{table}[!htb]
    
    \centering
    \begin{tabular}{l|c|c|>{\centering\arraybackslash}p{1.5cm}|>{\centering\arraybackslash}p{1.5cm}}
    %\toprule
        \multirow{2}{*}{Methods}    & Batch size    & \multirow{2}{*}{FPS}              & \multicolumn{2}{c}{Total training time (h)}\\
        \cmidrule{4-5}
                                    & per GPU       &                     & S4                    & MS3\\
        \midrule
        SAMA-AVS                                     & 2                 & 8             & 60.2                  & 18.8\\
        \rowcolor{LightGreen}SAVE (1024)               & 2                 & 8             & 60.2                  & 18.8\\
        \rowcolor{LightGreen}SAVE (256)               & \textbf{16}       & \textbf{35}   & \textbf{20.7}         & \textbf{6.9}\\
         %\bottomrule
    \end{tabular}
    \caption{Training batch size, time, and inference FPS of SAVE and SAMA-AVS on MS3 and S4 subsets.}
    \label{tab:speed_time}
\end{table}

\subsection{Ablation for audio encoder adapter}
In this section, we perform the study for choosing the number of layers in MLP of an audio encoder adapter.
The results are recorded in Table \ref{tab:layer_compare}.

 \begin{table}[!htb]
    
    \centering
    \begin{tabular}{l|c|c|c|c}
    %\toprule
        Subsets & \multicolumn{2}{c|}{S4 (V1S)}  &  \multicolumn{2}{c}{MS3 (V1M)} \\
        \midrule
        Methods & $\mathcal{M_{J}}$ & $\mathcal{M_{F}}$ & $\mathcal{M_{J}}$ & $\mathcal{M_{F}}$ \\
        \midrule
        2-layer (256)      & 81.68 & 0.887 & 62.94 & 0.696 \\
        3-layer (256)      & 81.58 & 0.880 & 62.54 & 0.686 \\
        \midrule
        2-layer (1024)      & 83.11 & 0.895 & 67.01 & 0.739 \\
        3-layer (1024)      & 83.01 & 0.891 & 66.91 & 0.719 \\
    %\bottomrule
    \end{tabular}
    \caption{Comparison of the number of layers of MLP.}
    \label{tab:layer_compare}
\end{table}

As we can see, using 2-layer MLP gives better results for both resolutions that we tested (256 and 1024). 
Thus, for the rest of our experiments, all of the MLP is a 2-layer MLP.
\section{Conclusion}
% The advent of large-scale pre-trained models has significantly improved the generalization capabilities of conventional computer vision tasks.
% However, generalizing cross-modal Audio-Visual Segmentation (AVS) in zero-shot and few-shot scenarios has received less attention.
In this study, we present SAVE, a potent audio-visual segmentation model that adheres to the encoder-prompt-decoder paradigm.
This model addresses the growing need for precise localization in real-world scenarios where annotated data is limited.
Our proposed method excels at achieving generalizable cross-modal segmentation.
It generates high-quality segmentation masks using only low-resolution input images.
Furthermore, SAVE facilitates faster training and serves as a memory-efficient framework, delivering high performance.
The extensive experiments show that SAVE outperforms the previous SOTA by a large margin across benchmark scenarios, from fully supervised to zero-shot and few-shot ones.

\bibliography{aaai25}

\clearpage
\appendix
\section{Qualitative Examples}
\begin{figure*}
    \centering
    \includegraphics[width=\linewidth]{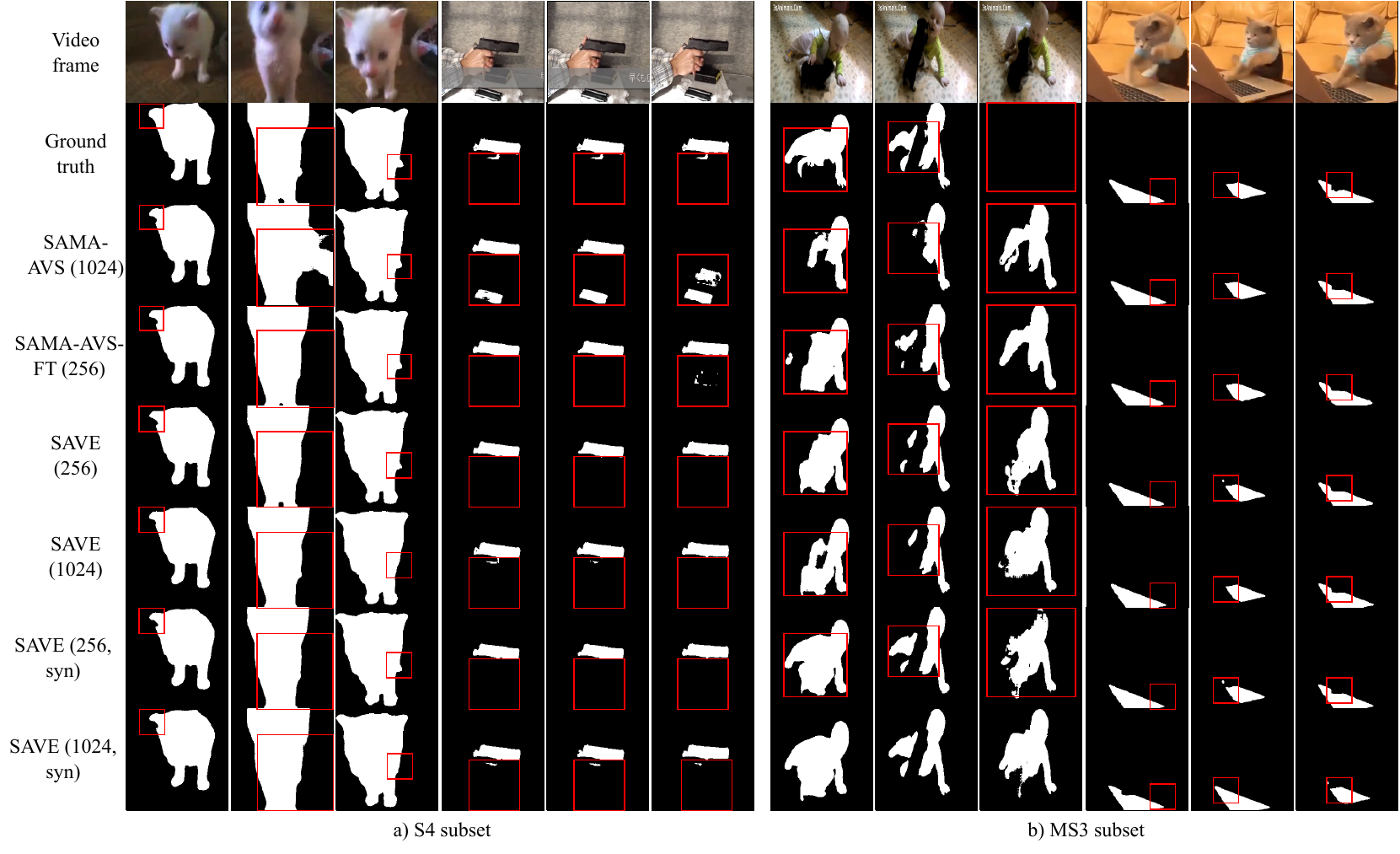}
    \caption{Qualitative results produced by our model and other methods on the test sets of S4 and MS3 subsets.}
    \label{fig:visual_comparison}
\end{figure*}
Figure \ref{fig:visual_comparison} presents a qualitative comparison of the results generated by SAMA-AVS and our proposed SAVE on AVSBench.
The results demonstrate that the segmentation masks produced by SAVE align more closely with the ground truths across both subsets.
Our model is capable of accurately segmenting the entire body of the sounding objects without omitting any parts.
It also outperforms SAMA-AVS in capturing the edges and intricate details of the sounding target, as evidenced by the clear depiction of the body of a cat and the gun ammo in Figure \ref{fig:visual_comparison}a.

Interestingly, SAMA-AVS struggled to differentiate between the sounding object and the associated object, such as the gun and the ammo.
In contrast, SAVE accomplished this task without generating incorrect masks.
In the MS3 setting, our model also demonstrated its ability to localize and segment multiple targets from the sound accurately.
Even in scenarios where multiple objects overlap or are absent in the ground truth, SAVE provides accurate segmentation masks, as illustrated in Figure \ref{fig:visual_comparison}b.

\end{document}